\documentclass[reprint,superscriptaddress,aps,prx]{revtex4-2}

\usepackage[usenames,dvipsnames]{xcolor}
\usepackage[utf8]{inputenc}
\usepackage{bm}
\usepackage{amsfonts}
\usepackage{graphicx,psfrag}

\usepackage{booktabs}
\usepackage{amsmath}
\usepackage{calc}
\usepackage{makecell}

\usepackage{nicefrac}

\usepackage{enumitem}
\usepackage{float}
\usepackage{natbib}
\usepackage{tikz}
\usepackage{soul,xcolor}
\usepackage[colorlinks=true,linkcolor=Blue,citecolor=Blue,urlcolor=Blue]{hyperref}

\usepackage{xfp}

\usepackage[caption=false]{subfig}
\usepackage[capitalise]{cleveref}

\newcommand{\revision}{\color{black}}

\newcommand{\x}{\mathbf{x}}
\newcommand{\f}{\mathbf{f}}
\newcommand{\g}{\mathbf{g}}

\newcommand{\G}{\mathbf{G}}
\newcommand{\y}{\mathbf{y}}

\newcommand{\W}{\mathbf{W}}

\newcommand{\Y}{\mathbf{Y}}
\newcommand{\xt}{\tilde{x}}
\newcommand{\yt}{\tilde{y}}
\newcommand{\A}{\mathbf{A}}
\newcommand{\B}{\mathbf{B}}
\def\*#1{\mathbf{#1}}

\newcommand{\norm}[1]{\left\lVert#1\right\rVert}

\newcommand{\Ntrain}{N_\textnormal{train}}
\newcommand{\Ntraj}{N_\textnormal{traj}}

\newtheorem{theorem}{Theorem}[section]
\newtheorem{result}[theorem]{Result}

\usepackage{xr}
\makeatletter

\newcommand*{\addFileDependency}[1]{
\typeout{(#1)}
\@addtofilelist{#1}
\IfFileExists{#1}{}{\typeout{No file #1.}}
}\makeatother

\newcommand*{\myexternaldocument}[1]{%
\externaldocument{#1}%
\addFileDependency{#1.tex}%
\addFileDependency{#1.aux}%
}

\myexternaldocument{Supp_material}
\begin{document}

\title{How more data can hurt: \\ Instability and regularization in next-generation reservoir computing}

\author{Yuanzhao Zhang}
\email{yzhang@santafe.edu}
\affiliation{Santa Fe Institute, 1399 Hyde Park Road, Santa Fe, NM 87501, USA}

\author{Edmilson Roque dos Santos}
\email{edmilson.roque.usp@gmail.com}
\affiliation{Department of Electrical and Computer Engineering, Clarkson University,
Potsdam, NY 13699, USA}
\affiliation{Clarkson Center for Complex Systems Science, Clarkson University,
Potsdam, NY 13699, USA}

\author{Huixin Zhang}
\email{huixinzhang@shu.edu.cn}
\affiliation{Department of Physics, Toronto Metropolitan University, Toronto, ON, M5B 2K3, Canada}
\affiliation{Department of Automation, Shanghai University, Shanghai 200444, China}

\author{Sean P. Cornelius}
\email{cornelius@torontomu.ca}
\affiliation{Department of Physics, Toronto Metropolitan University, Toronto, ON, M5B 2K3, Canada}

\begin{abstract} 
    It has been found recently that more data can, counter-intuitively, hurt the performance of deep neural networks. Here, we show that a more extreme version of the phenomenon occurs in data-driven models of dynamical systems. To elucidate the underlying mechanism, we focus on next-generation reservoir computing (NGRC)---a popular framework for learning dynamics from data. We find that, despite learning a better representation of the flow map with more training data, NGRC can adopt an ill-conditioned ``integrator'' and lose stability. We link this data-induced instability to the auxiliary dimensions created by the delayed states in NGRC. Based on these findings, we propose simple strategies to mitigate the instability, either by increasing regularization strength in tandem with data size, or by carefully introducing noise during training. Our results highlight the importance of proper regularization in data-driven modeling of dynamical systems.
\end{abstract}

\maketitle

\textbf{
It is not well understood how the amount and quality of data might affect the stability of machine-learning models in the context of learning unknown dynamics. This is different from typical tasks considered by the machine learning community such as regression or classification. Indeed, for forecasting tasks, the model often needs to be run in an auto-regressive fashion, which introduces the problem of stability (i.e., the prediction can be accurate in the short term but then fails catastrophically as errors accumulate over time). Here, we show that, surprisingly, more data (or higher quality data) can often hurt the performance of machine-learning models by inducing instability in their autonomous prediction mode.
We perform detailed analyses of this phenomenon using next-generation reservoir computing (NGRC)---a popular framework introduced recently for data-driven modeling of dynamical systems.  Using techniques from dynamical systems and numerical analysis, we pinpoint the mechanism behind data-induced instability to the auxiliary dimensions created by the delayed states in NGRC. 
Based on this finding, we propose two simple strategies to mitigate the instability. 
One is to increase the regularization strength in tandem with the amount of data, while the other is to intentionally decrease the quality of the data by a careful introduction of noise. This approach has the advantage that a fixed noise level can work for any amount of data.}

\section{Introduction}

Machine learning of dynamical systems (MLDS) has received increasing attention in recent years due to both its theoretical interest \cite{weinan2017proposal,levine2022framework,goring2024out,gilpin2024generative} and potential for wide-ranging applications from creating digital twins \cite{niederer2021scaling} to modeling climate \cite{li2020fourier} and controlling biological systems \cite{gilpin2020learning}.
Specifically, recent advances in MLDS have opened new possibilities in inferring effective brain connections~\cite{delabays2025hypergraph}, predicting tipping points \cite{kong2021machine}, discovering pattern-forming dynamics \cite{nicolaou2023data}, reconstructing chaotic attractors \cite{gilpin2020deep}, and anticipating synchronization transitions \cite{fan2021anticipating}. 

MLDS frameworks span the full spectrum from domain-agnostic deep-learning models to physics-informed models emphasizing interpretability and generalizability \cite{gilpin2023model}.
Domain-agnostic models are usually based on artificial neural networks or other ``universal approximator'' functions.
Popular examples include neural ODEs \cite{chen2018neural}, neural operators \cite{azizzadenesheli2024neural}, and reservoir computing \cite{pathak2018model}.
In contrast, physics-based models adopt strong inductive biases, potentially yielding greater accuracy at the expense of expressivity.
Some common paradigms in this space include symbolic regression \cite{Schmidt_2009} and  Sparse Identification of Nonlinear Dynamics (SINDy) \cite{brunton2016discovering}. 
Finally, there are also hybrid models \cite{pathak2018hybrid,karniadakis2021physics,chepuri2024hybridizing} that try to find the optimal balance between model scale and domain knowledge. 

It is natural to believe that---as in other machine learning tasks---more data is generally better in MLDS.
{\revision (A notable exception is the double descent phenomenon \cite{belkin2019reconciling,nakkiran2021deep}, which we discuss more in \cref{sec:discussion}.)}
Indeed, one major challenge faced by MLDS methods in many applications is a paucity of high-quality data (for example, due to noisy, sparse, and/or partial measurements).
Here, we show that sometimes too much high-quality data can also be a problem if one is not careful.
Specifically, we demonstrate that adding more (noise-free) data can hurt the performance of machine-learning models without commensurate regularization.
This counter-intuitive phenomenon stems from the potential long-term instability one must contend with in forecasting tasks.
In contrast to more ``static'' tasks such as image classification, simply fitting the model to a high-dimensional surface or distribution can be insufficient for data-driven models of dynamical systems---we also need to ensure the long-term stability of the model when run autonomously. 

We demonstrate the main phenomenon using next-generation reservoir computing (NGRC) \cite{gauthier2021next,bollt2021explaining}, which draws inspiration from the reservoir computing paradigm \cite{maass2002real,jaeger2004harnessing} but is more closely related to statistical forecasting methods such as nonlinear vector-autoregression (NVAR) \cite{billings2013nonlinear,jaurigue2022connecting}.
NGRC has been shown to perform well in many challenging forecasting and control tasks \cite{gauthier2022learning,barbosa2022learning,kent2024controlling,kent2024controlling1}, 
especially if appropriate nonlinear features are available \cite{zhang2023catch}.
The relatively simple structure of NGRC models allows us to systematically probe the mechanism behind data-induced instability.

\section{Results}
\label{sec:results}

\subsection{Next-generation reservoir computing}
\label{sec:ngrc}

Consider a dynamical system whose $n$-dimensional state $\mathbf{x}$ obeys a set of $n$ autonomous differential equations of the form
\begin{equation}
\dot{\x} = \f(\x).
\label{eq:continuous-dyn}
\end{equation}
NGRC aims to learn a representation of $\f$ through a discrete map of the form
\begin{equation}
\x_{t+1} = \x_t + \W \cdot \g_t,
\label{eq:ngrc-dyn}
\end{equation}
where the index $t$ runs over a set of discrete times separated by $\Delta t$, which represents the time resolution of the training data. $\W$ is the $n \times m$ readout matrix of trainable weights and
\begin{equation}
\g_t = \g\left(\x_t, \x_{t-1}, \ldots , \x_{t-k+1} \right)
\label{eq:features}
\end{equation}
is an $m$-dimensional vector of (non)linear features, calculated from the current state and $k - 1$ past states. Here, $k \ge 1$ is a hyperparameter that governs the amount of memory in the NGRC model.
The features generally include a constant (bias) term, $nk$ linear terms from the $k$ states, plus a number of user-specified nonlinear features. 

During training, given a time series $\{\x_t\}_{t=1,\cdots,\Ntrain}$, we seek a $\W$ that minimizes the least-square error between $\y_t$ and $\W \cdot \g_t$,
where $\y_t = \x_{t+1} - \x_t$. 
This is typically achieved via Ridge regression with Tikhonov regularization---a convex optimization problem with a unique solution given by
\begin{equation}
\W = \Y \G^T \left(\G \G^T + \lambda \mathbb{I}\right)^{-1}.
\label{eq:ridge-regression}
\end{equation}
Here $\Y$ ($\G$) is a matrix whose columns are the $\y_t$ ($\g_t$) and $\lambda \ge 0$ is the regularization coefficient. Note that training can be performed simultaneously on $\Ntraj > 1$ trajectories. In this case, one simply concatenates the regressors ($\g_t$) and regressands ($\y_t$) obtained from each trajectory.

During the prediction phase, we are given an initial condition $\x_0$ and $k - 1$ previous states $\x_{-1}$ to $\x_{1-k}$, with which we can iterate Eqs.~(\ref{eq:ngrc-dyn})-(\ref{eq:features}) as an autonomous dynamical system (each output becomes part of the model's input at the next time step). 
All simulations in this study are performed in Julia, using a custom implementation of NGRC. 

\subsection{Data-induced instability}
\label{sec:instability}

\begin{figure}[b]
\centering
\includegraphics[width=.99\linewidth]{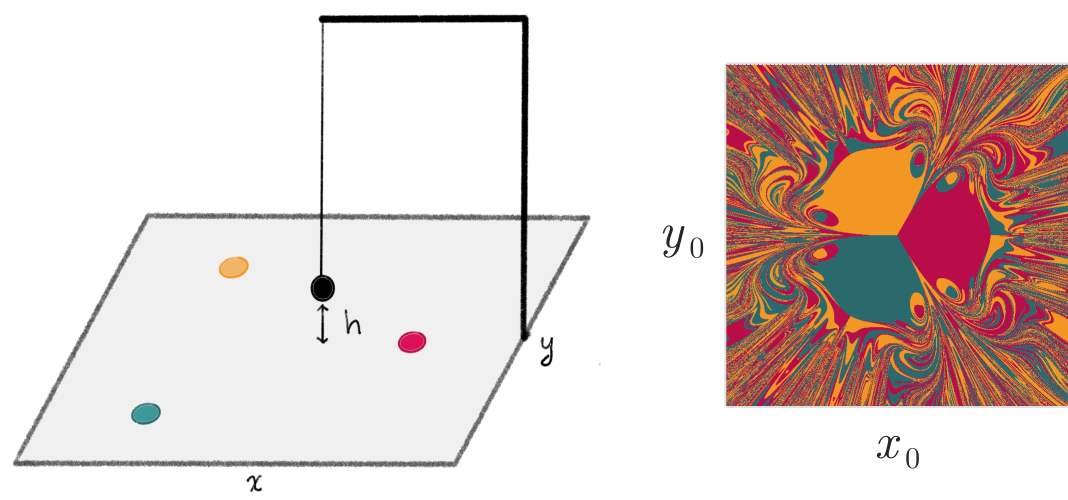}
\caption{
\textbf{Magnetic pendulum with three fixed-point attractors and corresponding basins of attraction}. (Left) Schematic of a magnetic pendulum system with three magnets. We take the coordinates of the magnets to be $\left(\nicefrac{1}{\sqrt{3}}, 0\right)$,\, $\left(\nicefrac{-1}{2 \sqrt{3}}, \nicefrac{-1}{2}\right)$,\, and $\left(\nicefrac{-1}{2\sqrt{3}}, \nicefrac{1}{2}\right)$ in dimensionless units. The $(x, y)$ coordinates of the pendulum bob and the corresponding velocities $(\dot{x}, \dot{y})$ fully specify the system's state. (Right) Simulated basins of attraction for the region of initial conditions under study, namely states of zero initial velocity with $-1.5 \le x_0, y_0 \le 1.5$.}
\label{fig:schematic}
\end{figure}

We first demonstrate the main phenomenon by applying NGRC to a representative nonlinear system---the magnetic pendulum \cite{motter2013doubly}.
The system consists of a ferromagnetic bob suspended above the origin of the  $(x, y)$ plane. There, three identical point magnets are placed at the vertices of an equilateral triangle with unit side length (\cref{fig:schematic}). The pendulum bob moves under the influence of the magnetic forces as well as gravity and frictional damping, resulting in the following equations of motion:
\begin{align} 
    \ddot{x} &= -\omega_0^2 x - a\dot{x} - \sum_{i=1}^3 \frac{x-\xt_i}{\left(\left(x-\xt_i\right)^2+\left(y-\yt_i\right)^2+h^2\right)^\frac{3}{2}} \label{eq:x-dyn}, \\
    \ddot{y} &= -\omega_0^2 y - a\dot{y} - \sum_{i=1}^3 \frac{y-\yt_i}{\left(\left(x-\xt_i\right)^2+\left(y-\yt_i\right)^2+h^2\right)^\frac{3}{2}} \label{eq:y-dyn}.
\end{align}
Here $(\xt_i,\yt_i)$ are the coordinates of the $i$th magnet, $\omega_0$ is the pendulum's natural frequency, $a$ is the damping coefficient, and $h$ is the bob's height above the plane. We focus on parameter values for which the system is multistable, with a total of three fixed-point attractors (one for each magnet). 

To learn the dynamics of the magnetic pendulum, we train NGRC on varying numbers of trajectories ($\Ntraj$) obtained from initial conditions with the bob at rest in different positions $(x_0, y_0)$. We then ask the trained model to reconstruct the basins of attraction on the $(x_0, y_0)$ plane, again with $\dot{x}_0 = \dot{y}_0 = 0$. That is, to accurately predict the final attractor for a wide range of different initial conditions. This is a challenging task that is crucial in many applications of multistable dynamical systems \cite{wiley2006size,delabays2017size,zhang2021basins,zhang2023deeper}.
We equip NGRC with $x-$ and $y-$ components of the magnetic force terms as its nonlinear features. 
In this case, NGRC is capable of learning the dynamics well enough to accurately reconstruct the basins \cite{zhang2023catch}. 
Later we will show that our results also hold for more generic nonlinearities, such as radial basis functions.

\begin{figure}[tb]
\centering
\includegraphics[width=1\columnwidth]{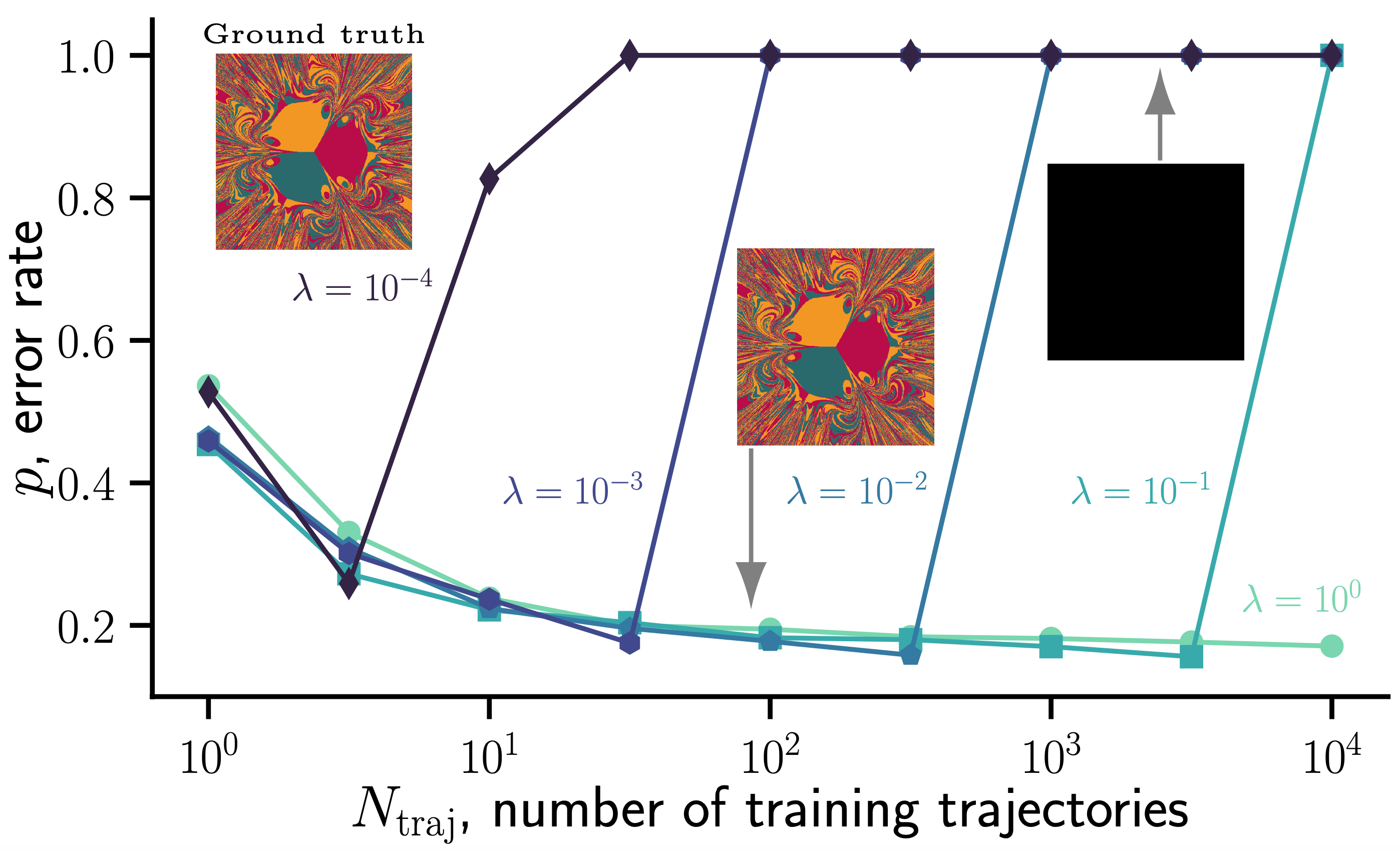}
\caption{
\textbf{More (noise-free) data can induce instability in NGRC.}
As we increase the number of trajectories $\Ntraj$ used to train NGRC (with pendulum force terms as nonlinear features), the model can undergo a sudden transition from accurately capturing the three fractal-like basins to losing stability and sending all trajectories to infinity.
This happens for a wide range of regularization coefficients $\lambda$, although larger $\lambda$ can delay the onset of instability.
We measure the performance of a trained NGRC model using the error rate $p$, which is the percentage of initial conditions for which the model predicts the wrong attractor.
Each data point is obtained by averaging $p$ over $10$ independent trials.
The ground-truth basins and representative NGRC predictions are shown as insets. Black denotes trajectories that diverged to infinity. 
Here, the pendulum parameters are set to $\omega_0 = 0.5$, $a = 0.2$, $h=0.2$, and the NGRC hyperparameters used are $k=2$, $\Delta t = 0.01$, and $\Ntrain = 3000$.
}
\label{fig:ngrc-instability}
\end{figure}

\Cref{fig:ngrc-instability} shows our first main result.
We see that as more training trajectories are included, NGRC models capture the intricate basins better and better, as expected.
However, at some point, the trained model becomes unstable---\emph{all} forecasted trajectories diverge to infinity instead of converging to one of the three fixed-point attractors.
At any fixed regularization strength ($\lambda$) this instability transition is attributable solely to the amount of training data. Indeed, once the number of training trajectories crosses a threshold, the NGRC model---perfectly stable when trained with less data---blows up.
This sudden breakdown occurs for a wide range of regularization coefficients $\lambda$, with more data needed to induce instability for more aggressive regularization.
This is surprising in light of the usual expectation that in machine learning, more data usually translates into better performance with less overfitting. Especially when the data are independent and noise-free, as here. {\revision Interestingly, the same phenomenon afflicts forecasting chaos in the Lorenz system, where larger regularization is needed to ensure model stability as $\Ntraj$ increases (Supplementary Material, Fig.~S1).}

So what is causing this data-induced instability?
We will offer two complementary explanations in the sections that follow.
But first, we demonstrate what is \emph{not} the cause, which happens to be most people's first guess.

\subsection{It is not the flow map!}
\label{sec:flow-map}

From \cref{eq:ngrc-dyn}, we see that NGRC learns the dynamics from data by constructing a map from $\x_t$ to $\x_{t+1}$.
In dynamical systems theory, a \emph{flow map} predicts the state of the system $\Delta t$ time units later based on the current state.
For example, the magnetic pendulum system induces a flow map $\Phi_{\Delta t}$ from $\mathbb{R}^4$ to $\mathbb{R}^4$: 
\begin{equation}\label{eq:flow_map}
    \Phi_{\Delta t}(x_t, y_t, \dot{x}_t, \dot{y}_t)=(x_{t+\Delta t}, y_{t+\Delta t}, \dot{x}_{t+\Delta t}, \dot{y}_{t+\Delta t}).
\end{equation}
Equivalently, we can encode the information in the flow surface:
\begin{equation}\label{eq:flow_surface}
    \phi_{\Delta t}(\x_t)=\Phi_{\Delta t}(\x_t) - \x_t=(\Delta x_t, \Delta y_t, \Delta \dot{x}_t, \Delta \dot{y}_t).
\end{equation}
Once an accurate flow surface has been constructed, NGRC can in principle make reliable forecasts from any initial conditions (at least in parts of the state space where there was enough training data).
In this sense, the goal of NGRC is not dissimilar to many other machine learning frameworks---find an accurate fit of the target high-dimensional manifold and use it to make predictions.

Of course, in practice, we only have a finite amount of data.
And like many other machine-learning tasks, one needs to worry about overfitting.
Could it be that as NGRC tries to fit more and more data points on the flow surface, it generates a pathological surface that contorts itself to fit the training data, diverging from the true flow surface in the gaps between?

\begin{figure}[tb]
\centering
\includegraphics[width=1\columnwidth]{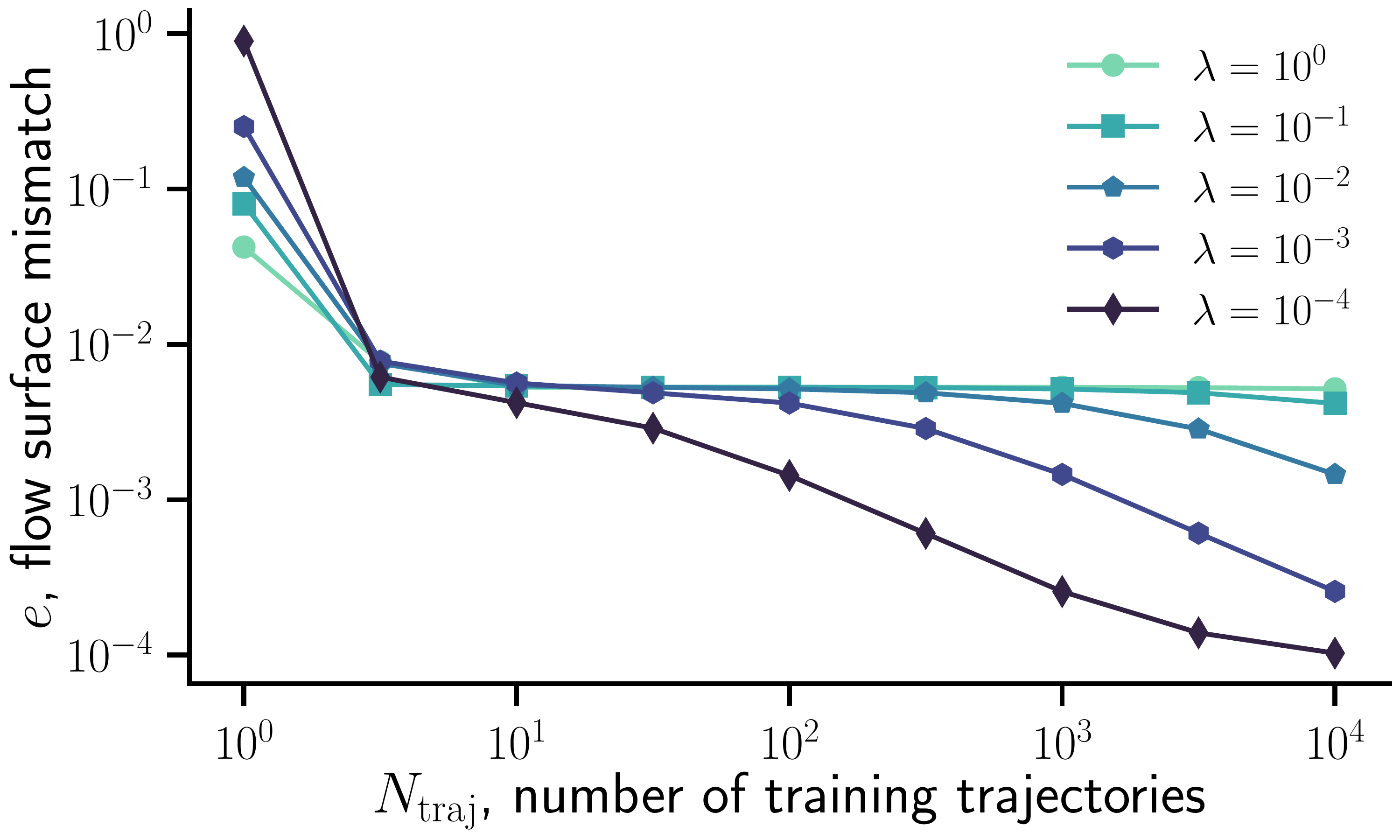}
\caption{
\textbf{NGRC does not overfit the flow surface.}
Using the same setup as in \cref{fig:ngrc-instability}, we plot the fitting error ($e$, \cref{eq:fitting-error})
against the number of training trajectories $\Ntraj$.
The fitting error is monotonically decreasing for all values of regularization coefficient $\lambda$ considered, giving no indication of any over-fitting that might lead to instability.
}
\label{fig:flow-map}
\end{figure}

\Cref{fig:flow-map} shows that this hypothesis, however natural, fails to explain the data-induced instability.
There, we characterize the mismatch between the model-predicted and true flow surfaces on a 2D sub-manifold $(x, y, 0, 0) \rightarrow \Delta \dot{x}$ via the fitting error ($e$) between the real/NGRC flow maps for the $x$-velocity. Namely,
\begin{equation}
e = \lVert \Delta \dot{x}^{\text{(real)}} - \Delta \dot{x}^{\text{(NGRC)}} \rVert.
\label{eq:fitting-error}
\end{equation}
We observe no signs of overfitting. In fact, as more and more training trajectories are included, $e$ keeps decreasing for all values of regularization coefficient $\lambda$ considered.
We emphasize that $e$ is evaluated on a uniform and dense grid independent of the training data points.
These results remain qualitatively the same if we examine the fitting error over other 2D sub-manifolds of the flow surface, such as $(0, 0, \dot{x}, \dot{y}) \rightarrow \Delta y$. 

\begin{figure*}[tb]
\centering
\includegraphics[width=1.6\columnwidth]{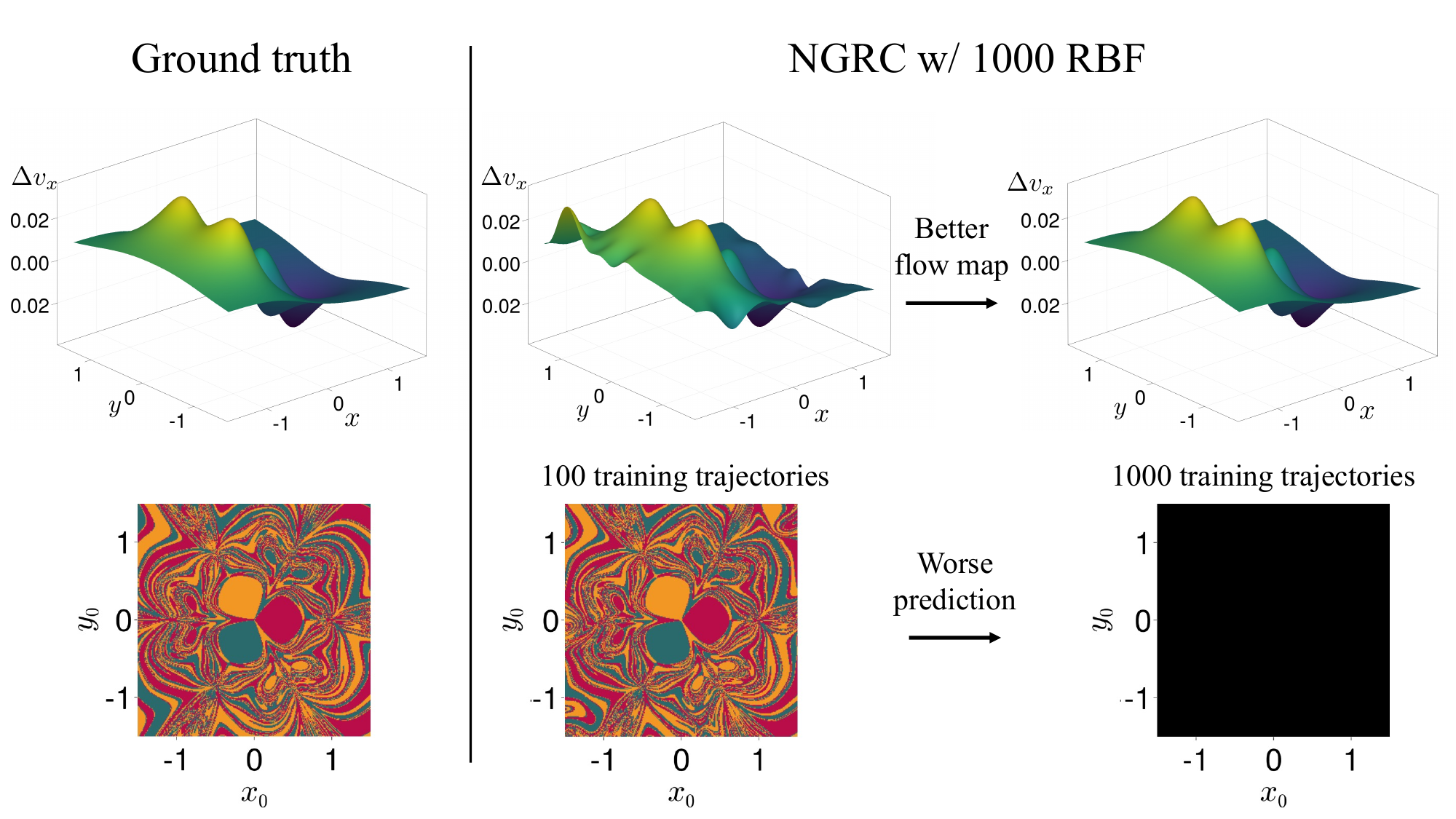}
\caption{
\textbf{Instability in NGRC cannot be explained by errors in fitting the flow surface.}
Here we demonstrate this point using the magnetic pendulum system and NGRC with 1000 radial basis functions (RBF) as nonlinear features.
Each RBF has a center chosen randomly and uniformly from $(x_0,y_0) \in [-1.5,1.5]^2$.
When trained with $\Ntraj=100$ trajectories, NGRC achieves a passable (but far from perfect) fit of the flow surface and does a good job in capturing the basins (error rate $p\approx 0.2$).
When we increased $\Ntraj$ to $1000$, the model learned a near-perfect flow map, but it became unstable and all predictions blew up.
The pendulum parameters are set to $\omega_0 = 0.5$, $a = 0.2$, $h=0.4$, and the NGRC parameters used are $k=2$, $\lambda=10^{-5}$, $\Delta t = 0.01$, and $\Ntrain = 5000$.
}
\label{fig:flow-map-rbf}
\end{figure*}

\Cref{fig:flow-map-rbf} further rules out overfitting as the cause of data-induced instability.
This time, we replace the pendulum force terms with 1000 radial basis functions (RBF) \cite{lowe1988multivariable} of the same form considered in Ref.~\cite{zhang2023catch}, so as to demonstrate that the same instability occurs for NGRC models with generic nonlinearities.
We see that NGRC is stable when trained with $100$ trajectories and predicts the basins well, despite noticeable deviations in its fit to the flow surface.
When trained with $1000$ trajectories, NGRC reconstructs a much better flow surface.
However, its predictions now diverge to infinity for all initial conditions tested.
If it is not the flow map, then what is the cause of this instability?

\subsection{Integrator perspective}
\label{sec:integrator}

\begin{table*}[]
\centering

\hspace{-14.3cm} Two-step Adams–Bashforth

\begin{tabular}{c|rrrrrrrrrrrrrrrrrrrr}
 & $x$ & $y$ & $v_x$  & $v_y$  & $\hat{x}$ & $\hat{y}$ & $\hat{v}_x$  & $\hat{v}_y$  & \tiny $F_{1,x}$  & \tiny $F_{1,y}$  & \tiny $F_{2,x}$  & \tiny $F_{2,y}$ & \tiny $F_{3,x}$ & \tiny $F_{3,y}$ & \tiny $F_{1,\hat{x}}$ & \tiny $F_{1,\hat{y}}$ & \tiny $F_{2,\hat{x}}$ & \tiny $F_{2,\hat{y}}$  & \tiny $F_{3,\hat{x}}$  & \tiny $F_{3,\hat{y}}$ \\ \hline
$\Delta x/ \Delta t$  & \textcolor{gray}{0.00} & \textcolor{gray}{0.00}  & \textcolor{orange}{1.50}  & \textcolor{gray}{0.00} & \textcolor{gray}{0.00}  & \textcolor{gray}{0.00}  & \textcolor{orange}{-0.50}  & \textcolor{gray}{0.00}  & \textcolor{gray}{0.00}  & \textcolor{gray}{0.00}  & \textcolor{gray}{0.00} & \textcolor{gray}{0.00} & \textcolor{gray}{0.00}  & \textcolor{gray}{0.00}  & \textcolor{gray}{0.00} & \textcolor{gray}{0.00} & \textcolor{gray}{0.00}  & \textcolor{gray}{0.00}  & \textcolor{gray}{0.00} & \textcolor{gray}{0.00} \\
$\Delta y/ \Delta t$ & \textcolor{gray}{0.00}  & \textcolor{gray}{0.00} & \textcolor{gray}{0.00} & 1.50  & \textcolor{gray}{0.00}  & \textcolor{gray}{0.00}  & \textcolor{gray}{0.00}  & -0.50  & \textcolor{gray}{0.00} & \textcolor{gray}{0.00}  & \textcolor{gray}{0.00} & \textcolor{gray}{0.00}  & \textcolor{gray}{0.00} & \textcolor{gray}{0.00}  & \textcolor{gray}{0.00} & \textcolor{gray}{0.00}  & \textcolor{gray}{0.00} & \textcolor{gray}{0.00}  & \textcolor{gray}{0.00} & \textcolor{gray}{0.00}  \\
$\Delta v_x/ \Delta t$ & -0.375 & \textcolor{gray}{0.00}  & \textcolor{red}{-0.30}  & \textcolor{gray}{0.00}  & 0.125 & \textcolor{gray}{0.00}  & \textcolor{red}{0.10} & \textcolor{gray}{0.00}  & \textcolor{ForestGreen}{1.50} & \textcolor{gray}{0.00}  & \textcolor{ForestGreen}{1.50} & \textcolor{gray}{0.00} & \textcolor{ForestGreen}{1.50} & \textcolor{gray}{0.00}  & \textcolor{ForestGreen}{-0.50} & \textcolor{gray}{0.00} & \textcolor{ForestGreen}{-0.50} & \textcolor{gray}{0.00}  & \textcolor{ForestGreen}{-0.50} & \textcolor{gray}{0.00} \\
$\Delta v_y/ \Delta t$ & \textcolor{gray}{0.00}  & -0.375 & \textcolor{gray}{0.00}  & \textcolor{blue}{-0.30}  & \textcolor{gray}{0.00}  & 0.125 & \textcolor{gray}{0.00}   & \textcolor{blue}{0.10} & \textcolor{gray}{0.00} & 1.50 & \textcolor{gray}{0.00} & 1.50  & \textcolor{gray}{0.00} & 1.50 & \textcolor{gray}{0.00} & -0.50  & \textcolor{gray}{0.00} & -0.50 & \textcolor{gray}{0.00} & -0.50 
\end{tabular}

\vspace{5mm}

\hspace{-14.3cm} $\Ntraj=10$, $\kappa=5$, $p=0.23$

\begin{tabular}{c|rrrrrrrrrrrrrrrrrrrr}
& $x$ & $y$ & $v_x$  & $v_y$  & $\hat{x}$ & $\hat{y}$ & $\hat{v}_x$  & $\hat{v}_y$  & \tiny $F_{1,x}$  & \tiny $F_{1,y}$  & \tiny $F_{2,x}$  & \tiny $F_{2,y}$ & \tiny $F_{3,x}$ & \tiny $F_{3,y}$ & \tiny $F_{1,\hat{x}}$ & \tiny $F_{1,\hat{y}}$ & \tiny $F_{2,\hat{x}}$ & \tiny $F_{2,\hat{y}}$  & \tiny $F_{3,\hat{x}}$  & \tiny $F_{3,\hat{y}}$ \\ \hline
$\Delta x/ \Delta t$ & \textcolor{gray}{0.00} & \textcolor{gray}{0.00}  & \textcolor{orange}{0.50}  & \textcolor{gray}{0.00} & \textcolor{gray}{0.00}  & \textcolor{gray}{0.00}  & \textcolor{orange}{0.50}  & \textcolor{gray}{0.00}  & \textcolor{gray}{0.00}  & \textcolor{gray}{0.00}  & 0.01 & \textcolor{gray}{0.00} & \textcolor{gray}{0.00}  & \textcolor{gray}{0.00}  & 0.01 & \textcolor{gray}{0.00} & \textcolor{gray}{0.00}  & \textcolor{gray}{0.00}  & 0.01 & \textcolor{gray}{0.00} \\
$\Delta y/ \Delta t$ & \textcolor{gray}{0.00}  & \textcolor{gray}{0.00} & \textcolor{gray}{0.00} & 0.50  & \textcolor{gray}{0.00}  & \textcolor{gray}{0.00}  & \textcolor{gray}{0.00}  & 0.50  & \textcolor{gray}{0.00} & \textcolor{gray}{0.00}  & \textcolor{gray}{0.00} & 0.01  & \textcolor{gray}{0.00} & \textcolor{gray}{0.00}  & \textcolor{gray}{0.00} & 0.01  & \textcolor{gray}{0.00} & \textcolor{gray}{0.00}  & \textcolor{gray}{0.00} & 0.01  \\
$\Delta v_x/ \Delta t$ & -0.13 & \textcolor{gray}{0.00}  & \textcolor{red}{1.06}  & -0.12 & -0.13 & \textcolor{gray}{0.00}  & \textcolor{red}{-1.26} & 0.12  & \textcolor{ForestGreen}{-0.49} & \textcolor{gray}{0.00}  & \textcolor{ForestGreen}{1.50} & \textcolor{gray}{0.00} & \textcolor{ForestGreen}{-0.49} & \textcolor{gray}{0.00}  & \textcolor{ForestGreen}{1.50} & \textcolor{gray}{0.00} & \textcolor{ForestGreen}{-0.49} & \textcolor{gray}{0.00} & \textcolor{ForestGreen}{1.50} & \textcolor{gray}{0.00} \\
$\Delta v_y/ \Delta t$ & \textcolor{gray}{0.00}  & -0.13 & -0.11 & \textcolor{blue}{1.30}  & \textcolor{gray}{0.00}  & -0.13 & 0.11  & \textcolor{blue}{-1.50} & \textcolor{gray}{0.00} & -0.49 & \textcolor{gray}{0.00} & 1.50  & \textcolor{gray}{0.00} & -0.49 & \textcolor{gray}{0.00} & 1.50  & \textcolor{gray}{0.00} & -0.49 & \textcolor{gray}{0.00} & 1.50 
\end{tabular}

\vspace{5mm}

\hspace{-14cm} $\Ntraj=100$, $\kappa=27$, $p=0.18$

\begin{tabular}{c|rrrrrrrrrrrrrrrrrrrr}
& $x$ & $y$ & $v_x$  & $v_y$  & $\hat{x}$ & $\hat{y}$ & $\hat{v}_x$  & $\hat{v}_y$  & \tiny $F_{1,x}$  & \tiny $F_{1,y}$  & \tiny $F_{2,x}$  & \tiny $F_{2,y}$ & \tiny $F_{3,x}$ & \tiny $F_{3,y}$ & \tiny $F_{1,\hat{x}}$ & \tiny $F_{1,\hat{y}}$ & \tiny $F_{2,\hat{x}}$ & \tiny $F_{2,\hat{y}}$  & \tiny $F_{3,\hat{x}}$  & \tiny $F_{3,\hat{y}}$ \\ \hline
$\Delta x/ \Delta t$ & \textcolor{gray}{0.00} & \textcolor{gray}{0.00}  & 0.53  & \textcolor{gray}{0.00} & \textcolor{gray}{0.00}  & \textcolor{gray}{0.00}  & 0.47   & \textcolor{gray}{0.00}   & \textcolor{gray}{0.00}  & \textcolor{gray}{0.00} & 0.01  & \textcolor{gray}{0.00} & \textcolor{gray}{0.00}  & \textcolor{gray}{0.00} & 0.01  & \textcolor{gray}{0.00} & \textcolor{gray}{0.00}  & \textcolor{gray}{0.00} & 0.01  & \textcolor{gray}{0.00} \\
$\Delta y/ \Delta t$ & \textcolor{gray}{0.00}  & \textcolor{gray}{0.00} & \textcolor{gray}{0.00} & 0.53  & \textcolor{gray}{0.00}  & \textcolor{gray}{0.00}  & \textcolor{gray}{0.00}   & 0.47   & \textcolor{gray}{0.00} & \textcolor{gray}{0.00}  & \textcolor{gray}{0.00} & 0.01  & \textcolor{gray}{0.00} & \textcolor{gray}{0.00}  & \textcolor{gray}{0.00} & 0.01  & \textcolor{gray}{0.00} & \textcolor{gray}{0.00}  & \textcolor{gray}{0.00} & 0.01  \\
$\Delta v_x/ \Delta t$ & -0.14 & \textcolor{gray}{0.00}  & \textcolor{red}{13.77} & -0.39 & -0.14 & \textcolor{gray}{0.00}  & \textcolor{red}{-14.00} & 0.39   & -0.43 & \textcolor{gray}{0.00} & 1.56  & \textcolor{gray}{0.00} & -0.43 & \textcolor{gray}{0.00} & 1.56  & \textcolor{gray}{0.00} & -0.43 & \textcolor{gray}{0.00} & 1.56  & \textcolor{gray}{0.00} \\
$\Delta v_y/ \Delta t$ & \textcolor{gray}{0.00}  & -0.14 & -0.40 & \textcolor{blue}{13.93} & \textcolor{gray}{0.00}  & -0.15 & 0.40   & \textcolor{blue}{-14.16} & \textcolor{gray}{0.00} & -0.43 & \textcolor{gray}{0.00} & 1.56  & \textcolor{gray}{0.00} & -0.43 & \textcolor{gray}{0.00} & 1.56  & \textcolor{gray}{0.00} & -0.43 & \textcolor{gray}{0.00} & 1.56 
\end{tabular}

\vspace{5mm}

\hspace{-14cm} $\Ntraj=1000$, $\kappa=215$, $p=1$

\begin{tabular}{c|rrrrrrrrrrrrrrrrrrrr}
 & $x$ & $y$ & $v_x$  & $v_y$  & $\hat{x}$ & $\hat{y}$ & $\hat{v}_x$  & $\hat{v}_y$  & \tiny $F_{1,x}$  & \tiny $F_{1,y}$  & \tiny $F_{2,x}$  & \tiny $F_{2,y}$ & \tiny $F_{3,x}$ & \tiny $F_{3,y}$ & \tiny $F_{1,\hat{x}}$ & \tiny $F_{1,\hat{y}}$ & \tiny $F_{2,\hat{x}}$ & \tiny $F_{2,\hat{y}}$  & \tiny $F_{3,\hat{x}}$  & \tiny $F_{3,\hat{y}}$ \\ \hline
$\Delta x/ \Delta t$ & \textcolor{gray}{0.00} & \textcolor{gray}{0.00} & 0.72   & \textcolor{gray}{0.00}   & \textcolor{gray}{0.00}  & \textcolor{gray}{0.00} & 0.28    & \textcolor{gray}{0.00}   & \textcolor{gray}{0.00} & \textcolor{gray}{0.00} & 0.01 & \textcolor{gray}{0.00} & \textcolor{gray}{0.00} & \textcolor{gray}{0.00} & 0.01 & \textcolor{gray}{0.00} & \textcolor{gray}{0.00} & \textcolor{gray}{0.00} & 0.01 & \textcolor{gray}{0.00} \\
$\Delta y/ \Delta t$ & \textcolor{gray}{0.00} & \textcolor{gray}{0.00} & \textcolor{gray}{0.00}   & 0.72   & \textcolor{gray}{0.00} & \textcolor{gray}{0.00}  & \textcolor{gray}{0.00}   & 0.28    & \textcolor{gray}{0.00} & \textcolor{gray}{0.00} & \textcolor{gray}{0.00} & 0.01 & \textcolor{gray}{0.00} & \textcolor{gray}{0.00} & \textcolor{gray}{0.00} & 0.01 & \textcolor{gray}{0.00} & \textcolor{gray}{0.00} & \textcolor{gray}{0.00} & 0.01 \\
$\Delta v_x/ \Delta t$ & -0.26 & \textcolor{gray}{0.00} & \textcolor{red}{108.68} & 0.51 & -0.26 & \textcolor{gray}{0.00} & \textcolor{red}{-109.11} & -0.51   & 0.05 & \textcolor{gray}{0.00} & 2.03 & \textcolor{gray}{0.00} & 0.05 & \textcolor{gray}{0.00} & 2.03 & \textcolor{gray}{0.00} & 0.05 & \textcolor{gray}{0.00} & 2.03 & \textcolor{gray}{0.00} \\
$\Delta v_y/ \Delta t$ & \textcolor{gray}{0.00} & -0.26 & 0.42 & \textcolor{blue}{109.56} & \textcolor{gray}{0.00} & -0.26 & -0.42 & \textcolor{blue}{-109.98} & \textcolor{gray}{0.00} & 0.05 & \textcolor{gray}{0.00} & 2.04 & \textcolor{gray}{0.00} & 0.05 & \textcolor{gray}{0.00} & 2.04 & \textcolor{gray}{0.00} & 0.05 & \textcolor{gray}{0.00} & 2.04
\end{tabular}
\caption{
\textbf{NGRC learns an increasingly unstable integrator as more data are included in the training.}
Here, we compare NGRC models with the closest numerical integrator---two-step Adams-Bashforth from the linear multistep methods family.
We use the same setup to train the NGRC models as in \cref{fig:ngrc-instability}, with $\lambda=0.01$ and $\Ntraj$ ranging from $10$ to $1000$.
The NGRC models are stable for $\Ntraj=10$ and $100$ (error rate $p=0.23$ and $0.18$, respectively), but it becomes unstable for $\Ntraj=1000$ ($p=1$).
To save space, we denote the delayed states with a hat (e.g., $\hat{x}$) and the $x-$ and $y-$ components of the nonlinear terms with $F_{i,\cdot}$.
The key observation is that, as $\Ntraj$ is increased, we have an increasingly severe imbalance between the weights assigned to current states and delayed states.
For the two-step Adams-Bashforth method, the current state is assigned a weight that is $\frac{3}{2}$ times the corresponding coefficient on the right-hand side of the ODE, whereas the delayed state has a factor of $-\frac{1}{2}$.
This assignment is optimal in the sense that the factors $\{\frac{3}{2},-\frac{1}{2}\}$ are the only pair that makes the method an order-two integrator (i.e., its local truncation error is $\mathcal{O}(\Delta t^3)$).
In contrast, the learned weights are significantly larger and become increasingly unbalanced as more training data are used.
For example, see the pairs highlighted in red and blue for models trained with different $\Ntraj$.
The factors are almost equal numbers with opposite signs and, despite being orders of magnitude larger, their sum largely cancels each other and roughly match the corresponding coefficient on the right-hand side of the ODE.
This imbalance is directly reflected in the increasingly large condition number $\kappa$ of the NGRC readout matrix and is what eventually leads to the data-induced instability.
}
\label{tbl:integrator}
\end{table*}

In this subsection, inspired by the original paper \cite{gauthier2021next}, we view NGRC models as integrators, and we show that the NGRC instability can be understood using techniques from numerical analysis.
For the purpose of demonstration, it is easiest to consider NGRC models with pendulum force terms as nonlinear features. In this case, all terms in the original ODE [\cref{eq:x-dyn,eq:y-dyn}] are available in NGRC's feature library. This allows direct comparison of NGRC models with traditional integration schemes.

The integrators that most resemble the NGRC architecture are the linear multistep methods, defined by
\begin{equation}
    \sum_{j=0}^s a_j \mathbf{x}_{i+j}=\Delta t \sum_{j=0}^s b_j \mathbf{f}\left(t_{i+j}, \mathbf{x}_{i+j}\right),
\end{equation}
which uses the states and derivatives from $s$ previous steps to predict $\mathbf{x}_{i+s}$ and integrate the equation $\dot{\mathbf{x}} = \mathbf{f}(t, \mathbf{x})$ forward.
(For implicit methods, such as Adams–Moulton, $b_s\neq 0$ and the integrator also uses information about the derivative at the future step.) 
Because NGRC models make updates using only the current and past states [c.f.~\cref{eq:ngrc-dyn}], it is most analogous to the popular Adams–Bashforth methods, which are explicit methods with $a_{s-1}=-1$, $a_{s-2}=\cdots=a_0=0$, and $b_s=0$.
The other $b_j$ ($0\leq j < s$) are chosen such that the method has order $s$ (\emph{i.e.}, the local truncation error of the integrator is of order $\mathcal{O}(\Delta t^{s+1})$), which determines the $b_j$ uniquely.
We stress, however, that an NGRC model with exact nonlinearities is not necessarily a special case of Adams–Bashforth or similar methods. This is because it works directly with different scalar features (instead of the vector field $\mathbf{f}$ as a whole).
Accordingly, NGRC models can assign different weights to individual terms in its features library, unconstrained by the right-hand side of the ODEs.

For $k=2$, the natural integrator to compare NGRC models with is the two-step Adams–Bashforth method:
\begin{equation}
    \mathbf{x}_{i+2}=\mathbf{x}_{i+1}+ \Delta t \left( \frac{3}{2} \mathbf{f}\left(t_{i+1}, \mathbf{x}_{i+1}\right)-\frac{1}{2} \mathbf{f}\left(t_i, \mathbf{x}_i\right) \right).
\end{equation}
Do NGRC models approximately rediscover the two-step Adams–Bashforth method on their own?
In \cref{tbl:integrator}, we compare the readout matrix of NGRC models 
with the two-step Adams–Bashforth method.
For this purpose, it is helpful to rewrite \cref{eq:x-dyn,eq:y-dyn} as first-order ODEs:
\begin{align}
& \dot{x}=v_x, \label{eq:pendulum-1}\\
& \dot{y}=v_y, \label{eq:pendulum-2}\\
& \dot{v}_x=-\omega_0^2 x-a v_x+\sum_{i=1}^3 F_{i,x}, \label{eq:pendulum-3}\\
& \dot{v}_y=-\omega_0^2 y-a v_y+\sum_{i=1}^3 F_{i,y},
\label{eq:pendulum-4}
\end{align}
where we use $F_{i,\cdot}$ denotes the $x-$ and $y-$ components of the force from the $i$th magnet.
From \cref{eq:pendulum-1,eq:pendulum-2,eq:pendulum-3,eq:pendulum-4}, we can easily infer what two-step Adams–Bashforth thinks each entry in the readout matrix should be, which are summarized in the top table (for $\omega_0=0.5$ and $a=0.2$). 
Now looking at the trained NGRC models, we see they are quite different from the two-step Adams–Bashforth method.
For example, an inspection of the weights for $\Ntraj=10$ reveals that instead of splitting the coefficients between the current state and the delayed state $\frac{3}{2}$ to $-\frac{1}{2}$ (optimal, as in Adams–Bashforth), NGRC splits the coefficients equally for many of the linear terms (compare terms highlighted in orange).
The weights do capture the $\frac{3}{2}$ to $-\frac{1}{2}$ split for the nonlinear terms.
However, the model is often confused about which way the split should go, routinely assigning the $-\frac{1}{2}$ factor to the current state instead of the delayed state (compare terms highlighted in green).

The most striking departure from Adams–Bashforth, however, lies in the coefficients for $v_x$ and $v_y$ in relation to \cref{eq:pendulum-3,eq:pendulum-4}, which are highlighted in red and blue, respectively.
There, instead of the optimal factor pairs $\{\frac{3}{2},-\frac{1}{2}\}$, NGRC models assign factors far away from zero to both the current state and the delayed state.
These factors are similar in magnitude but opposite in signs and, when combined together, roughly recover the corresponding coefficient on the right-hand side of the ODE (i.e., the sum of the factors is close to $1$).
These magnitudes grow rapidly with $\Ntraj$, eventually making the NGRC model unstable for $\Ntraj=1000$.

At this point, one might say, wait a moment, isn't a flow map all you need to advance a dynamical system in time? Where does the integrator come in? If NGRC has a perfect fit of the flow surface, how can it still be unstable? In the next subsection, we address these questions by identifying a source of data-induced instability: the ill-conditioning of the matrix $\G^T$ in \cref{eq:ridge-regression}. Each row of $\G^T$ represents the evaluation of the various candidate features using a set of $k$ consecutive states from a training trajectory. For small $\Delta t$, when the same features are evaluated at successive time points, the corresponding column vectors in $\G^T$ are nearly linearly dependent, leading to a nearly rank-deficient $\G^T$. This ill-conditioning explains the uncontrolled growth of the NGRC weights with $\Ntraj$ observed in \cref{tbl:integrator}. 



\subsection{Ill-conditioned $\G^T$ is a source of data-induced instability}
\label{sec:geometric}

\begin{figure*}[tb]
\centering
\includegraphics[width=1.6\columnwidth]{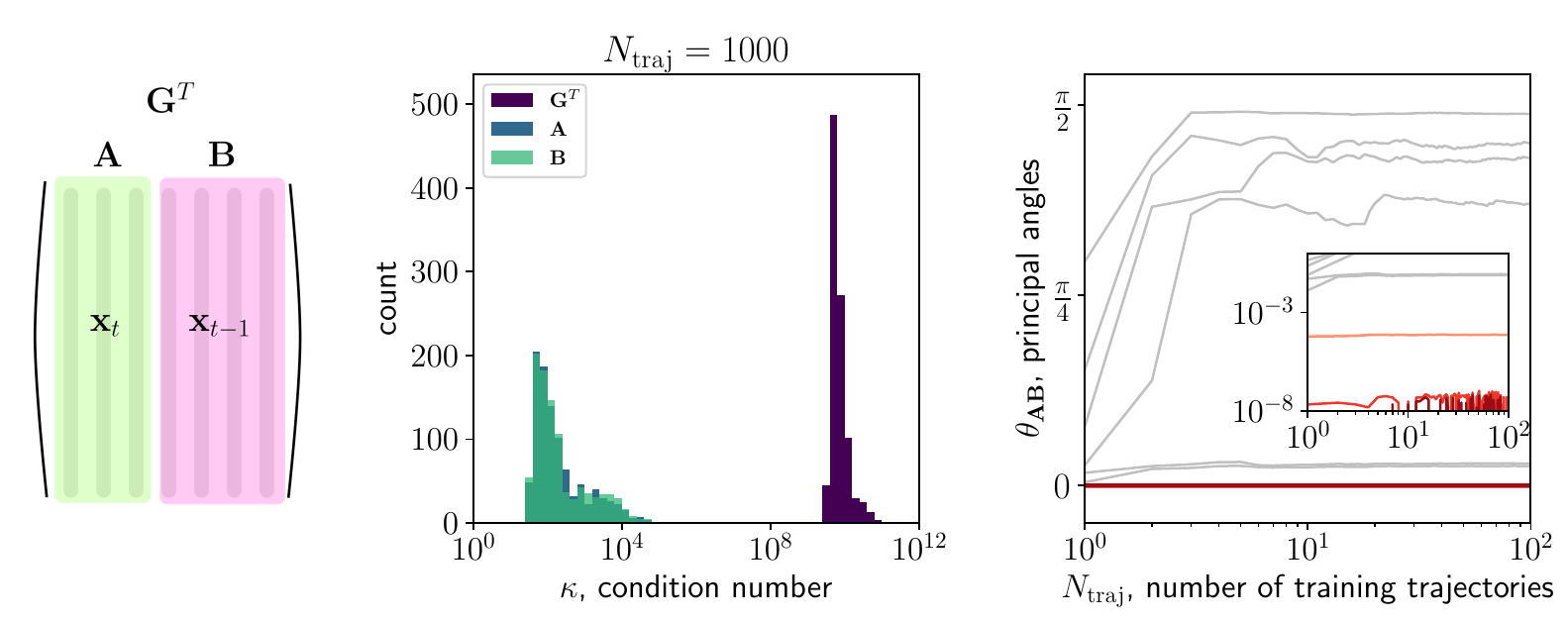}
\caption{
\textbf{Ill-conditioning arises from the linear dependence between current and delayed states.}
The left panel illustrates that for $k = 2$, the matrix $\G^T$ can be partitioned into two submatrices corresponding to the features evaluated at the current and delayed states along the time series. For the magnetic pendulum, $\A$ and $\B$ are $\Ntrain\Ntraj \times 10$ and $\Ntrain\Ntraj \times 11$ matrices, respectively. The middle panel displays the histogram of the condition number of each matrix $\G^T$, $\A$, and $\B$ for $\Ntraj = 1000$ trajectories. While $\A$ and $\B$ are well-conditioned, with overlapping histograms around $\mathcal{O}(10^{3})$, $\G^T$ is ill-conditioned, exhibiting a condition number orders of magnitude larger than its submatrices. The right panel shows the ten principal angles between the column spaces of $\A$ and $\B$ as the number of trajectories is increased. Even a hundredfold increase in trajectories has no impact on the group of principal angles near zero, indicating linear dependence between the column spaces $\mathcal{R}(\A)$ and $\mathcal{R}(\B)$. This is corroborated by the inset panel, which highlights a few principal angles below $10^{-1}$ on a logarithmic scale. Principal angles are color-coded: those below $10^{-3}$ appear in red, while larger angles are shown in gray. All settings and parameters are the same as in \cref{fig:ngrc-instability}. 
}
\label{fig:PABS}
\end{figure*}

To better understand the source of the ill-conditioning of $\G^T$, we use $k=2$ as an example and hence split $\G^T$ into two submatrices
\begin{align}
    \G^T = [\A, \B],
\end{align}
where $\A$ is the $\Ntrain \Ntraj \times l$ matrix with feature terms evaluated at the current time, and $\B$ is the $\Ntrain \Ntraj \times m$ matrix collecting all remaining terms, which include the bias term and the features evaluated at delayed states, as illustrated in the left panel of \cref{fig:PABS}. For the magnetic pendulum, the original equations of motion comprise $l = 10$ features in $\A$, while $\B$ includes the constant feature and delayed counterparts resulting in $m = 11$.
As shown in the middle panel in \cref{fig:PABS}, ill-conditioning of $\G^T$ cannot be attributed to either the submatrix $\A$ or $\B$ individually. We compute the condition numbers $\kappa(\G^T)$, $\kappa(\A)$, and $\kappa(\B)$ from one trajectory, and the histogram represents the distribution of these condition numbers over different trajectories. The matrix $\G^T$ is ill-conditioned when compared to its submatrices, with $\kappa(\G^T)$ typically around $\mathcal{O}(10^{10})$. 

\begin{figure}[tb]
\centering
\includegraphics[width=1\columnwidth]{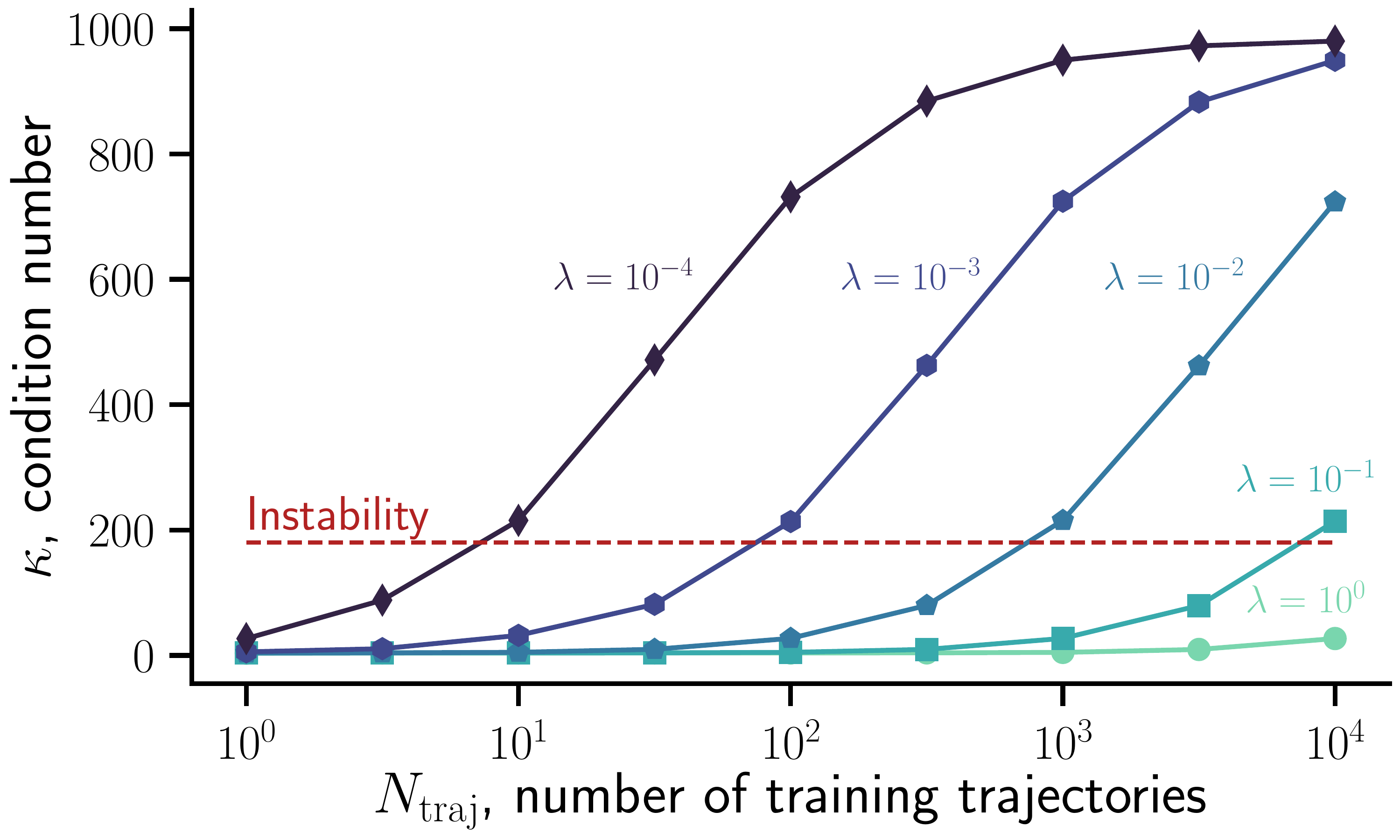}
\caption{
\textbf{Condition number as an indicator for instability in NGRC.}
Using the same setup as in \cref{fig:ngrc-instability}, we plot the condition number $\kappa(\W)$ of the readout matrix against the number of training trajectories $\Ntraj$.
The condition number is monotonically increasing as more training trajectories are included.
Once a threshold has been crossed ($\kappa(\W)\approx 180$), NGRC becomes unstable (c.f.~\cref{fig:ngrc-instability}).
We note, however, that this threshold based on the condition number is likely system-specific.
}
\label{fig:condition-number}
\end{figure}

What happens when more data is added? Adding different trajectories into the matrix $\G^T$ expands the space spanned by its column vectors. For example, due to the high sensitivity of the magnetic pendulum to initial conditions, trajectories sampled uniformly during training are likely to be distinct. While enlarging $\G^T$ would theoretically improve its conditioning, this is not observed numerically. Notably, the ill-conditioning persists as more training data is added. The right panel in \cref{fig:PABS} shows that the linear dependence between the column space of $\B$, $\mathcal{R}(\B)$, and of $\A$, $\mathcal{R}(\A)$, quantified by principal angles between subspaces (PABS) \cite{Bjorck_1973}, remains even with a hundredfold increase of number of trajectories. A cluster of principal angles $\theta_{\A\B}^i \in [0, \frac{\pi}{2}]$ for $i = 1, \dots, \min\{l, m\}$ near zero indicates that at least one direction in $\mathcal{R}(\A)$ aligns with $\mathcal{R}(\B)$, regardless of the amount of training data. This near-linear dependence persists because features are evaluated at consecutive times. Consequently, adding data only increases the size of $\G^T$ used in the Ridge regression (\cref{eq:ridge-regression}), leading to an increase in the norm of $\W$, or equivalently, the condition number $\kappa(\W)$, which tends to grow as shown in \Cref{fig:condition-number}. 

These numerical observations can be made more precisely by analyzing in more detail the least square problem ($\lambda = 0$ in \cref{eq:ridge-regression}). The least square problem approximates $\Y$, induced by the flow surface $\phi_{\Delta t}$ in \cref{eq:flow_surface}, with a vector lying in the column space of $\G^T$. Since $\G^T$ is a partitioned matrix, the column space of $\A$ corresponds to directions associated with feature terms in the original equations of motion \cref{eq:pendulum-1,eq:pendulum-2,eq:pendulum-3,eq:pendulum-4}, and thus the flow surface itself. In contrast, the transverse directions of the flow surface are spanned by vectors lying in $\mathcal{R}(\B)$. Therefore, it follows that
\begin{result}
    For $\lambda = 0$, whenever the minimum principal angle $\theta_{\A\B}$ is close to zero, the solution of \cref{eq:ridge-regression}, $\W$, has nonzero entries (corresponding to the transverse directions) that grow with $\Ntraj$.
\end{result}
The idea is to adapt Ref.~\cite{NOVAES2021132895} in the context of the NGRC model, using sensitivity analysis of the exact least-square regression problem ($\lambda = 0$ in \cref{eq:ridge-regression}). We refer the reader to SM.\ref{SM-sec:unstable_NGRC_transverse} in the Supplementary Material for the details. Ill-conditioning in the data matrix $\G$ and increasing the $\Ntraj$ inevitably leads to growing weights (entries of $\W$) corresponding to transverse directions in the solution of the least-squares problem \cite{GoluVanl96}. The result also extends to a small regularization coefficient $\lambda$, where $\W$ is computed using Ridge regression in \cref{eq:ridge-regression}. These growing coefficient entries of $\W$ in the transverse direction potentially lead to an unstable NGRC model. Without a mechanism to constrain the NGRC dynamics, NGRC does not learn a stable integration scheme, and the autonomous dynamics consequently diverge to infinity after a handful of iterations.


\begin{figure}[tb]
\centering
\includegraphics[width=1\columnwidth]{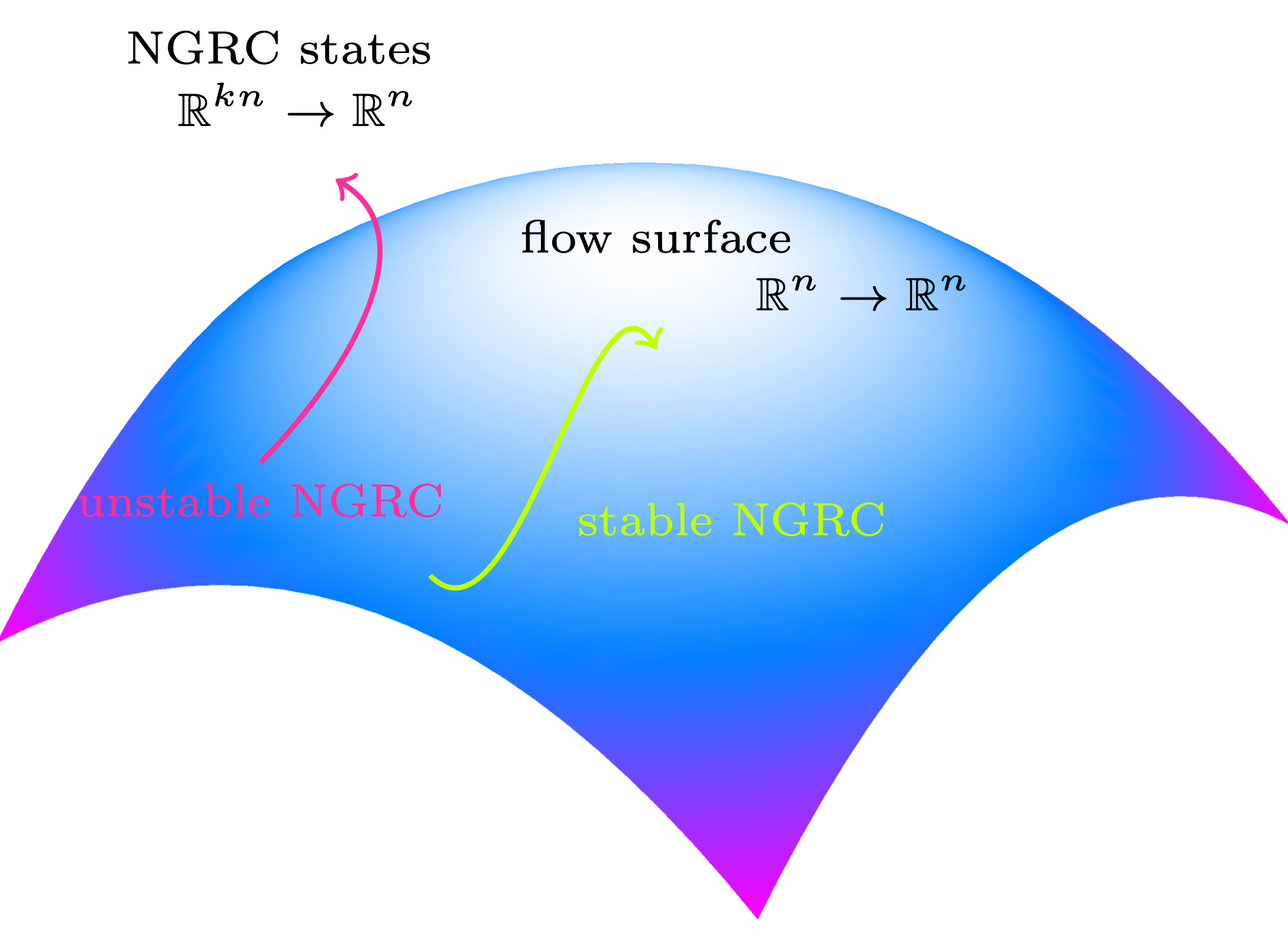}
\caption{
\textbf{Stable NGRCs stay close to the flow surface whereas unstable} NGRCs move away from the flow surface. During the prediction phase, a stable NGRC model would stay close to the low-dimensional sub-manifold where the $\mathbb{R}^n\rightarrow\mathbb{R}^n$ flow surface is defined.
This is also where all the training data come from (and thus where meaningful fitting happens).
An unstable NGRC model, in contrast, would move rapidly away from the flow surface due to transverse instability created by the delayed states, which typically results in the predictions blowing up.
}
\label{fig:cartoon}
\end{figure}

\begin{figure}[tb]
\centering
\includegraphics[width=1\columnwidth]{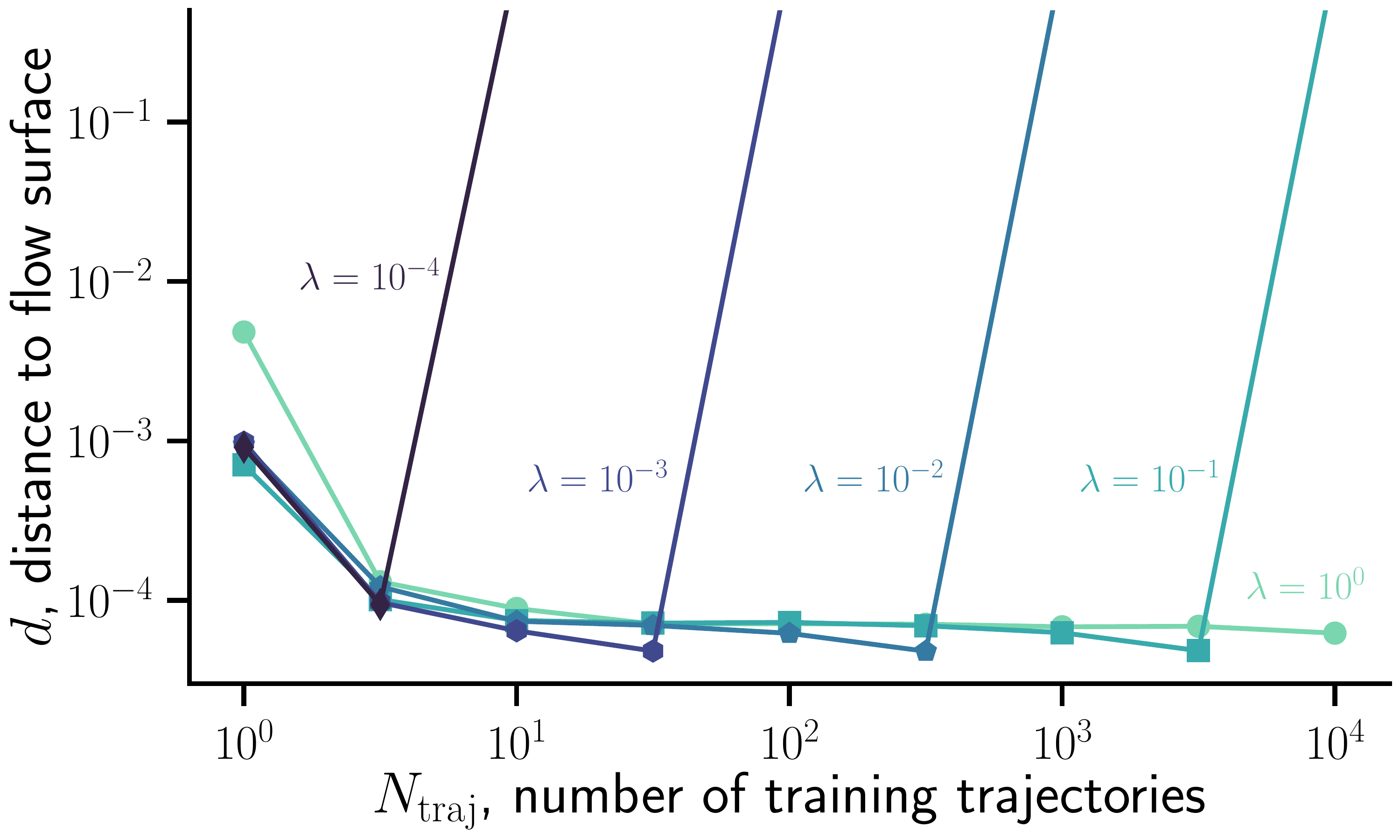}
\caption{
\textbf{More data can create transverse instability on the flow surface.}
Using the same setup as in \cref{fig:ngrc-instability}, we plot the average distance to the flow surface against the number of training trajectories.
The average distance $d$ is obtained by averaging the distance to the flow surface over the NGRC-predicted trajectory for $100$ time units. 
For a given regularization coefficient $\lambda$, transverse instability develops beyond a certain number of training trajectories, and the NGRC predictions diverge exponentially away from the low-dimensional flow surface ($d$ jumps from around $10^{-4}$ to numerical infinity).
We note that the transverse instability sets in exactly when NGRC becomes unstable (c.f. \cref{fig:ngrc-instability}).
}
\label{fig:distance_to_flow_surface}
\end{figure}

Geometrically, the key insight is the following: for the magnetic pendulum system, the flow map $\Phi_{\Delta t}$ is defined on $\mathbb{R}^4 \rightarrow \mathbb{R}^4$;
for NGRC, because of the inclusion of delayed states, the map $\Psi$ is actually defined on $\mathbb{R}^{4k} \rightarrow \mathbb{R}^4$. Thus, for $k > 1$, NGRC is learning a higher dimensional map than the true flow map. During training, because all data come from the real system, NGRC only sees the $\mathbb{R}^4 \rightarrow \mathbb{R}^4$ sub-manifold of $\Psi$, on which $\Phi_{\Delta t}$ is defined. 
It is thus no surprise that, as NGRC tries to fit more and more data on the $\mathbb{R}^4 \rightarrow \mathbb{R}^4$ sub-manifold, it inevitably creates instability in the other $4(k-1)$ directions transverse to the sub-manifold.
When such transverse instabilities exist, NGRC trajectories will move away exponentially from that sub-manifold (so as long as we do not start exactly on the subspace where the flow map is defined).
In this case, the fitting error on the flow surface is meaningless, no matter how good the fit is.
The instability of NGRC (as an integrator) we characterized in \cref{sec:integrator} is hidden and not reflected in the flow surfaces visualized in \cref{sec:flow-map} because we are using the ground truth for the delayed states.
Instead, the instability only becomes apparent when looking at the full $\mathbb{R}^{4k} \rightarrow \mathbb{R}^4$ space (in particular, at directions transverse to the low-dimensional flow surface).
We illustrate this geometric origin of NGRC instability in \cref{fig:cartoon} and show the transverse instability explicitly in \cref{fig:distance_to_flow_surface}.

The results above have a few interesting corollaries. First, we can make $\G^T$ better-conditioned (and the corresponding NGRC model more stable) simply by setting $k = 1$ (\emph{i.e.}, no delayed states). The nearly linear dependence between the current and delayed state is absent, making the entries of $\W$ not grow with more training data. Indeed, we no longer observe any instability for NGRC in this case. However, we pay a price by eschewing the delayed states: although stable, NGRC becomes much less expressive---it cannot fit the flow surface well and the prediction performance is poor unless the sampling rate is unrealistically high (\emph{i.e.}, $\Delta t \to 0$). Alternatively, $\G^T$ becomes less ill-conditioned if the time skip (delay) between the current state and the delayed states is increased, see SM.\ref{SM-sec:time_skip} in the Supplementary Material for details. However, increasing the time skip demands more training data in addition to creating a new hyperparameter to optimize, which can be challenging in practical applications. Finally, the NGRC model can also be stabilized by reducing $\kappa(\W)$. 
For example, for a fixed $\Ntraj$, if $\kappa(\W)$ is above the instability threshold, we can stabilize NGRC by increasing the regularization coefficient $\lambda$. Operationally, more aggressive regularization helps to dampen the pseudo-inverse of $\G^T$, filtering components corresponding to small singular values and mitigating the spurious linearly dependent directions between the current and delayed states.

\subsection{How to fix this?}
\label{sec:fix}

The game we have to play is controlling the ill-conditioning of $\G^T$ to avoid the unstable NGRC dynamics. So are there simple ways to mitigate data-induced instabilities? Here we discuss two simple strategies to overcome the ill-conditioned $\G^T$: data size-dependent regularization coefficient and training with noise.

\noindent
\textbf{Data size-dependent regularization coefficient:} To make progress, it is helpful to recall the loss function for Ridge regression
\begin{equation}
    \ell(\W) = \sum_{i=1}^{\Ntraj}\norm{\Y_i - \W \cdot \G_i}^2 + \lambda \norm{\W}^2.
    \label{eq:L2}
\end{equation}
Here, $\norm{\cdot}$ is the Frobenius norm (elementwise $L_2$ norm).
$\Y_i$ ($\G_i$) is a matrix whose columns are the states $\y_t$ (features $\g_t$) from the $i$-th training trajectory.
Assuming that each training trajectory gives rise to a similar fitting error, we can see that the first term in \cref{eq:L2} is proportional to the number of training trajectories $\Ntraj$.
Thus, as increasing the training data, the fitting error in \cref{eq:L2} becomes more and more dominant, effectively decreasing the regularization strength $\lambda$ and eventually leading to an under-regularized model.
In order to maintain a roughly constant model complexity (measured here by $\norm{\W}$), the regularization coefficient $\lambda$ should scale linearly with the amount of training data. 


\begin{figure}[tb]
\centering
\includegraphics[width=1\columnwidth]{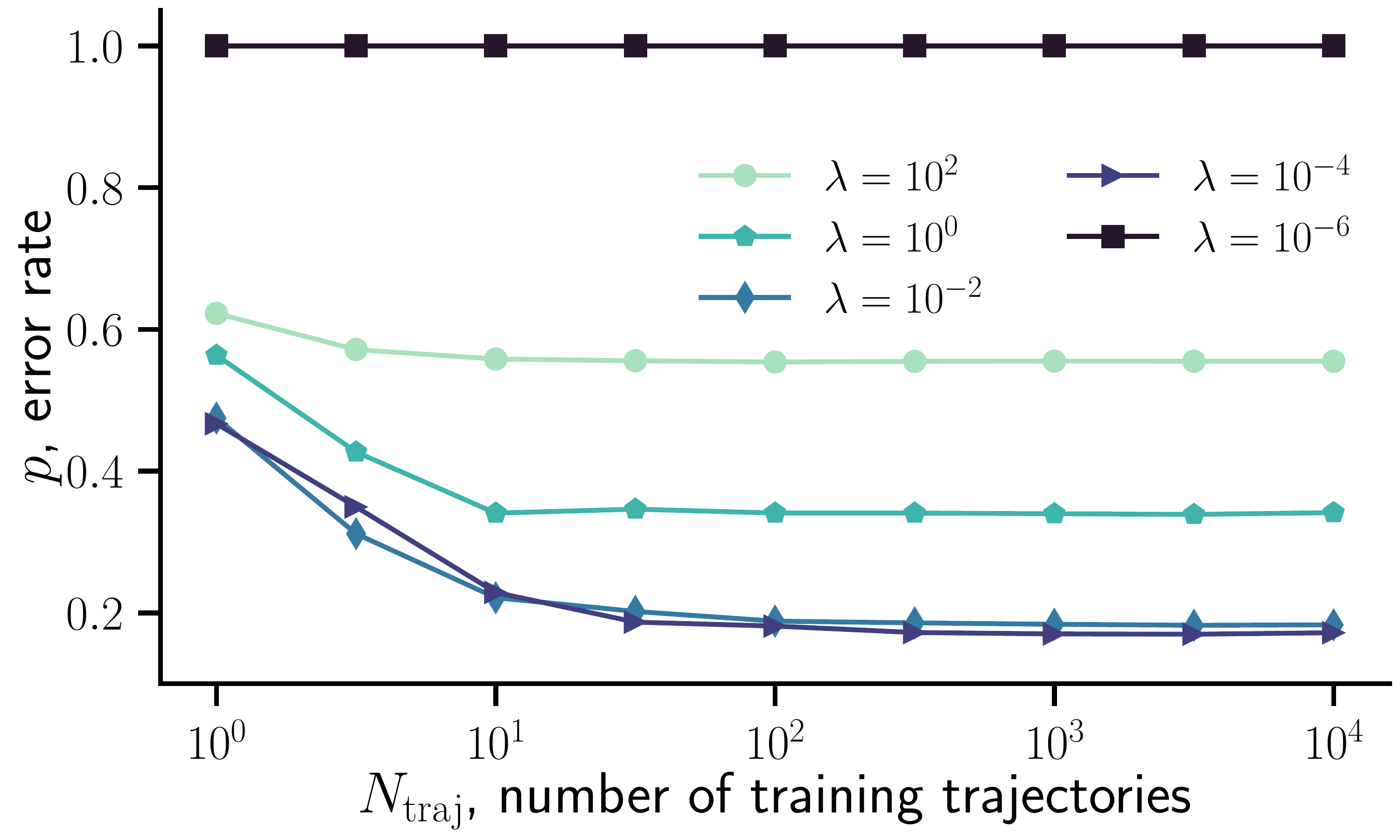}
\caption{
\textbf{Proper scaling of the regularization strength fixes data-induced instability in NGRC.}
This is the analog of \cref{fig:ngrc-instability} but with the regularization coefficient $\lambda$ scaled linearly by the amount of training data [c.f. \cref{eq:scaled-L2}].
Now as we increase the number of training trajectories $\Ntraj$, the trained NGRC no longer undergoes instability transitions, and the error rate $p$ stabilizes for large $\Ntraj$.
If the regularization is insufficient, e.g., $\lambda=10^{-6}$, then the trained NGRC is unstable for all $\Ntraj$.
If the regularization is too large, e.g., $\lambda=10^{2}$, it will negatively impact the prediction accuracy.
All settings and parameters other than the regularization are the same as in \cref{fig:ngrc-instability}.
}
\label{fig:ngrc-instability-with-scaled-regularization}
\end{figure}

Next, we revisit the magnetic pendulum system using the new loss function defined below, which increases regularization accordingly as more training data are introduced:
\begin{equation}
    \ell(\W) = \sum_{i=1}^{\Ntraj}\norm{\Y_i - \W \cdot \G_i}^2 + \lambda\Ntraj \norm{\W}^2.
    \label{eq:scaled-L2}
\end{equation}
\Cref{fig:ngrc-instability-with-scaled-regularization} shows the basin prediction error rate $p$ as a function of $\Ntraj$.
Compared to \cref{fig:ngrc-instability}, we see that there is no longer an instability transition at large $\Ntraj$, showing that the scaled regularization in \cref{eq:scaled-L2} can successfully suppress the data-induced instability.
We have also confirmed that the condition number $\kappa$ becomes roughly independent of $\Ntraj$, indicating a nearly constant model complexity $\norm{\W}$ regardless of the amount of training data.
As expected, there is a sweet spot of intermediate regularization that optimizes accuracy while ensuring stability.
In the case of \cref{fig:ngrc-instability-with-scaled-regularization}, it is $\lambda \in (10^{-4}, 10^{-1})$.

\begin{figure}[tb]
\centering
\includegraphics[width=1\columnwidth]{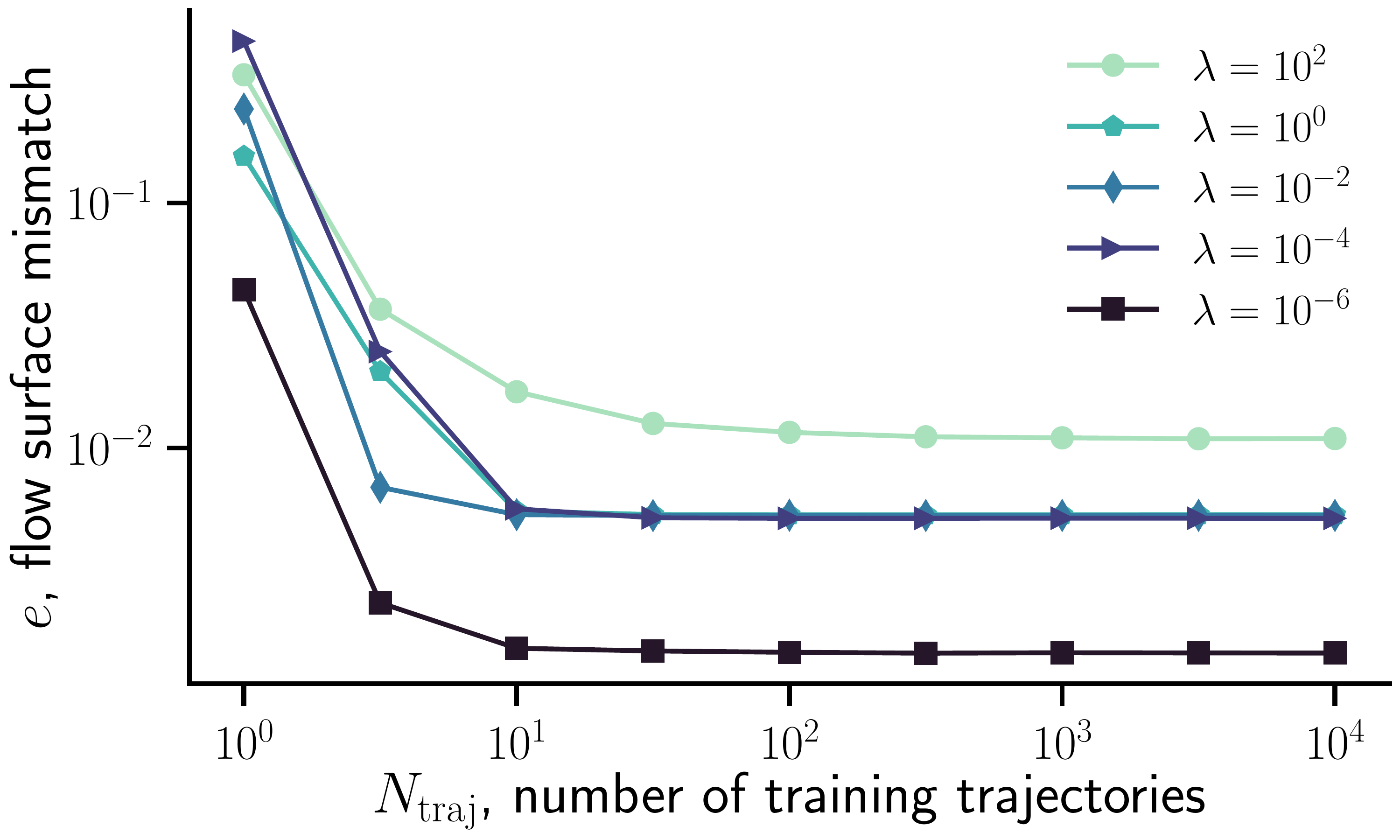}
\caption{
\textbf{Under scaled regularization, fitting error to the flow surface saturates as more training data are included.}
This is the analog of \cref{fig:flow-map} but with the regularization coefficient $\lambda$ scaled linearly by the amount of training data [c.f. \cref{eq:scaled-L2}].
Instead of the fitting error $e$ continuously decreasing with $\Ntraj$, it stabilizes quickly as the optimization achieves a balance between the error term and the regularization term.
All settings and parameters other than the regularization are the same as in \cref{fig:flow-map}.
}
\label{fig:flow-map-with-scaled-regularization}
\end{figure}

\Cref{fig:flow-map-with-scaled-regularization} tells a similar story: after an initial decrease in the fitting error of the flow surface, it approaches a constant value as $\Ntraj$ is further increased.
Smaller $\lambda$ leads to smaller $e$, but a $\lambda$ that is too small (e.g., $\lambda=10^{-6}$) triggers instability transverse to the flow surface and destabilizes the NGRC model.

\noindent
\textbf{Training with noise:} Aside from properly scaling the Ridge regression regularization strength $\lambda$, a possible alternative strategy to mitigate instability is adding noise to the training data. Such noise regularization has been shown to improve the performance and robustness of reservoir computers \cite{wikner2024stabilizing}.
There are also intimate connections between noise regularization and Ridge regression when i.i.d.\ noise is added directly to the regressors \cite{van2015lecture}.
One potential advantage of regularizing with noise is that a constant level of noise may suffice to stabilize the trained model independent of the amount of training data used.

We have tested this possibility by applying additive noise to the training time series before calculating the features $\mathbf{g}_t$ [c.f.~\cref{eq:features}]. 
At each time step $t$, we draw noise uniformly and independently from $\left[-\sigma, \sigma\right]$ for each of the $n \cdot k$ components in $\lbrace \mathbf{x}_t, \mathbf{x}_{t-1}, \ldots, \mathbf{x}_{t-k+1}\rbrace$. 
Here, $\sigma$ is a parameter that controls the noise strength. 
We stress that only the training inputs are modified; the next-step targets ($\mathbf{y}_t =\mathbf{x}_{t+1}-\mathbf{x}_t$) are still determined based on the noise-free data. Noise breaks the linear dependence between current and delayed states because random matrices whose entries are i.i.d. contain almost surely a set of linear independent random vectors. Intuitively, this forces the NGRC model to learn a contraction map that flows back to the flow surface in its vicinity. 

\begin{figure}[tb]
\centering
\includegraphics[width=1\columnwidth]{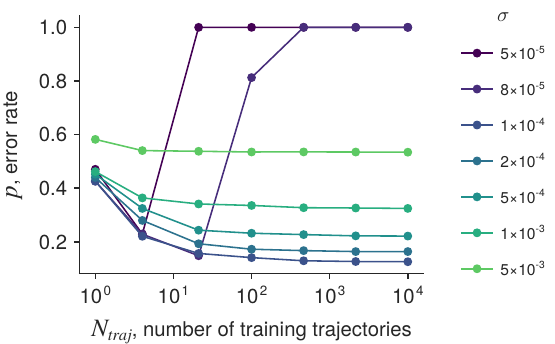}
\caption{\textbf{Noise-based regularization can stabilize NGRC but is sensitive to the choice of noise strength.} Unlike Tikhonov regularization in \cref{fig:ngrc-instability}, noise-based regularization with a fixed strength can stabilize NGRC models for all data sizes. However, the noise strength $\sigma$ needs to be chosen carefully to ensure stability without sacrificing accuracy. Each data point represents the average over 10 independent trials. Here, we set $\lambda = 10^{-4}$. All other hyperparameters are the same as in \cref{fig:ngrc-instability}.}
\label{fig:noise-regularization}.
\end{figure}

\Cref{fig:noise-regularization} shows the basin prediction error $p$ versus the number of training trajectories $\Ntraj$ for varying noise strengths $\sigma$. 
For this analysis, we set the (unscaled) Tikhonov regularization $\lambda = 10^{-4}$, for which model instability appeared at $\Ntraj \approx 10$ sans noise regularization (c.f.~\cref{fig:ngrc-instability}). 
We see that the effects of noise are a mixed blessing. 
With sufficient noise strength, it does mitigate the harmful effects of more data.
In particular, for a given noise strength $\sigma$, the basin prediction error ($p$) is asymptotically independent of $\Ntraj$. But the precise value of $p$ one attains as $\Ntraj \rightarrow \infty$ is exquisitely sensitive to the noise strength. 
When $\sigma$ is too low (e.g., $8 \times 10^{-5}$), noise fails to stabilize; we still see a transition to $p = 100\%$ at a finite value of $\Ntraj$. Increasing $\sigma$ only slightly to $10^{-4}$ stabilizes the model regardless of $\Ntraj$ and in fact achieves the lowest asymptotic error rate observed. 
Unfortunately, increasing $\sigma$ any further from here rapidly degrades model performance (\cref{fig:noise-regularization}). 
All told, there is a narrow range of $\sigma$---less than an order of magnitude---in which noise is strong enough to be an effective regularizer but not so strong that it hinders the learning of the flow map.


\section{Discussion}
\label{sec:discussion}

Here we have shown how more data can induce instability when learning unknown dynamical systems.
We focused on next-generation reservoir computing, a simple but powerful framework recently introduced to complement the standard reservoir computing paradigm \cite{gauthier2021next}.
We showed that the data-induced instability observed in NGRC does not come from overfitting the flow map.
Instead, it has to do with instabilities along auxiliary dimensions transverse to the flow map.
Finally, we linked such instabilities to the fact that training with increasing amounts of data at fixed regularization can lead to under-regularized models. 
Motivated by this observation, we proposed simple fixes either by increasing regularization strength in proportion to data size or by introducing the right amount of noise.

{\revision{
Similar instabilities can arise in other high-dimensional machine learning models that involve solving a linear regression with an ill-conditioned feature matrix. For instance, in network reconstruction problems, adding “unnecessary” features can result in ill-conditioning \cite{NOVAES2021132895}. Traditional reservoir computers (e.g., echo-state networks) \cite{pathak2017using,lu2018attractor,grigoryeva2018echo,carroll2019network,carroll2020reservoir,patel2021using,wikner2021using,rohm2021model,flynn2021multifunctionality,barbosa2021symmetry,kim2021teaching,gottwald2021combining,ma2023novel,gonon2023approximation,kong2023reservoir,kong2024reservoir,li2024higher,yan2024emerging} also exhibit ill-conditioned matrices  \cite{Dutoit_2009_regularization} that affect dynamical stability. For example, \cite{lu2018attractor} reports a failure to reproduce the correct Lyapunov exponents without regularization. However, the source of ill-conditioning in RCs differs from that in NGRC. Traditional RCs do not explicitly use past states of the system as features; instead, the dynamics emerge from the evolution of the reservoir states. The stability of these dynamics is typically controlled by the spectral radius of the reservoir’s internal weight matrix. Setting the spectral radius less than one ensures a contraction in the reservoir state space, which, for some classes of systems, guarantees the echo state property \cite{Grigoryeva_2019}. Moreover, hyperparameters like the spectral radius and regularizer parameter are often tuned in an optimization method to maintain stability during testing, hiding such data-induced instability. 
}}
This is different from NGRC, where the readout matrix needs to ``deal by itself'' with potential instabilities in the NGRC dynamics. Thus, the benefit of NGRC in reducing the computational cost of optimizing in higher dimensional hyper-parameter space requires a proper regularization scaling.

The deleterious effect of more training data is also a crucial aspect of the ``double descent'' phenomenon \cite{belkin2019reconciling,belkin2021fit}, in which model performance depends non-monotonically on the ratio of parameters/data. Double descent has been demonstrated in simple regression tasks \cite{nakkiran2019more} as well as deep neural networks such as ResNet and Transformers \cite{nakkiran2021deep}.
It has been shown that regularization plays a key role in the double descent phenomenon \cite{belkin2020two,d2020double,mei2022generalization,schaeffer2023double,davies2023unifying}.
In particular, double descent mostly appears in under-regularized models, and optimal regularization can often completely suppress the test error peak at the interpolation threshold (i.e., the model performance improves monotonically with model size and data size) \cite{nakkiran2020optimal}.
These similarities suggest a tantalizing connection between the results here and the double descent phenomenon.
However, we haven't observed a second descent in the context of basin prediction with NGRC (e.g., adding more data does not eventually stabilize NGRC).
In the future, it would be interesting to explore different model sizes and look for double-descent curves as more features are included in NGRC or bigger reservoirs are used in RC \cite{ribeiro2021beyond}.

\newpage

\begin{acknowledgments}
We thank Dan Gauthier and David Wolpert for pointing out the importance of normalizing $\lambda$ with data size and for providing references.
We also thank Michelle Girvan, William Gilpin, Brian Hunt, Zachary Nicolaou, and Anastasia Bizyaeva for insightful discussions.
Y.Z. acknowledges support from the Omidyar Fellowship and the National Science Foundation (NSF) under Grant No. DMS2436231. 
E.R.S. acknowledges support from the National Institute on Alcohol Abuse and Alcoholism under Grant No. R01AA029926. 
S.P.C. acknowledges support from the Natural Science and Engineering Research Council (NSERC, RGPIN-2020-05015) and the Digital Research Alliance of Canada.
\end{acknowledgments}


\section*{Code availability}
Our source code can be found at \url{https://github.com/spcornelius/RCBasins}.

\section*{Author contributions}
Y.Z. and S.P.C. designed the research and performed the initial study. E.R.S. developed the theory for data-induced instability in NGRC dynamics. H.Z. performed the numerical simulations of the forecasting task for the Lorenz system. All authors analyzed the data and wrote the paper.


\bibliography{bibli}

\end{document}